\documentclass[lettersize,journal]{IEEEtran}
\usepackage{amsmath,amsfonts}
\usepackage{algorithmic}
\usepackage{algorithm}
\usepackage{array}
\usepackage[caption=false,font=footnotesize,labelfont=sf,textfont=sf]{subfig}
\usepackage{textcomp}
\usepackage{stfloats}
\usepackage{url}
\usepackage{verbatim}
\usepackage{graphicx}
\usepackage{cite}

\usepackage{booktabs}
\usepackage{multirow}
\usepackage{multicol}
\usepackage{subfig}
\begin{document}

\title{Co-Occurrence Matters: Learning Action Relation for Temporal Action Localization}

\author{Congqi Cao\textsuperscript{* \textdagger},~\IEEEmembership{Member,~IEEE,},Yizhe Wang\textsuperscript{*}, Yue Lu, Xin Zhang, and Yanning Zhang,~\IEEEmembership{Senior Member,~IEEE}
	\thanks{\textsuperscript{*}Equal contribution. \textsuperscript{\textdagger}Corresponding author.}
	\thanks{This work is supported by the National Natural Science Foundation of China (Project No. U19B2037, 61906155), the Key R\&D Project of Shaanxi Province (Project No. 2023-YBGY-240), the Young Talent Fund of Association for Science and Technology in Shaanxi, China (Project No. 20220117), and the National Key R\&D Program of China (No. 2020AAA0106900).}
	\thanks{Congqi Cao, Yizhe Wang, Yue Lu, Xin Zhang, and Yanning Zhang are with the National Engineering Laboratory for Integrated Aero-Space-Ground-Ocean Big Data Application Technology, School of Computer Science, Northwestern Polytechnical University, Xi’an 710129, China (e-mail: congqi.cao@nwpu.edu.cn; wyz2927@163.com; zugexiaodui@gmail.com; zhangxin\_@mail.nwpu.edu.cn; ynzhang@nwpu.edu.cn).}
}


\markboth{Journal of \LaTeX\ Class Files,~Vol.~14, No.~8, August~2021}%
{Shell \MakeLowercase{\textit{et al.}}: A Sample Article Using IEEEtran.cls for IEEE Journals}

\IEEEpubid{0000--0000/00\$00.00~\copyright~2021 IEEE}

\maketitle

\begin{abstract}
Temporal action localization (TAL) is a prevailing task due to its great application potential.
Existing works in this field mainly suffer from two weaknesses: (1) They 
often neglect the multi-label case and only focus on temporal modeling. (2) They ignore the semantic information in class labels and only use the visual information.
To solve these problems, we propose a novel Co-Occurrence Relation Module (CORM) that explicitly models the co-occurrence relationship between actions.
Besides the visual information, it further utilizes the semantic embeddings of class labels to model the co-occurrence relationship. 
The CORM works in a plug-and-play manner and can be easily incorporated with the existing sequence models.
By considering both visual and semantic co-occurrence, our method achieves high multi-label relationship modeling capacity.
Meanwhile, existing datasets in TAL always focus on low-semantic atomic actions.
Thus we construct a challenging multi-label dataset UCF-Crime-TAL that focuses on high-semantic actions by annotating the UCF-Crime dataset at frame level and considering the semantic overlap of different events.
Extensive experiments on two commonly used TAL datasets, \textit{i.e.}, MultiTHUMOS and TSU, and our newly proposed UCF-Crime-TAL demenstrate the effectiveness of the proposed CORM, which achieves state-of-the-art performance on these datasets.
\end{abstract}


\section{Introduction}
\IEEEPARstart{A}{mong} the field of computer vision, temporal action localization (TAL) \cite{lin2018bsn,liu2019multi,tirupattur2021modeling,9660459,9454500,8440749,DBLP:conf/ijcai/YangW22,DBLP:journals/tcsv/SunSYX23,DBLP:conf/bmvc/KazakosHNZD21,DBLP:conf/cvpr/Zhang0Z21,Zhou_2023_WACV,Nag_2023_WACV,8741082,10058582,DBLP:conf/bmvc/DaiDB21} has received lots of attention due to its great potential in real-world applications, such as security monitoring, human-computer interaction, \textit{etc}.
Given an untrimmed video, TAL aims to localize where the actions occur and identify their categories.

\begin{figure}[!t]
	\centering
	\includegraphics[width=3.5in]{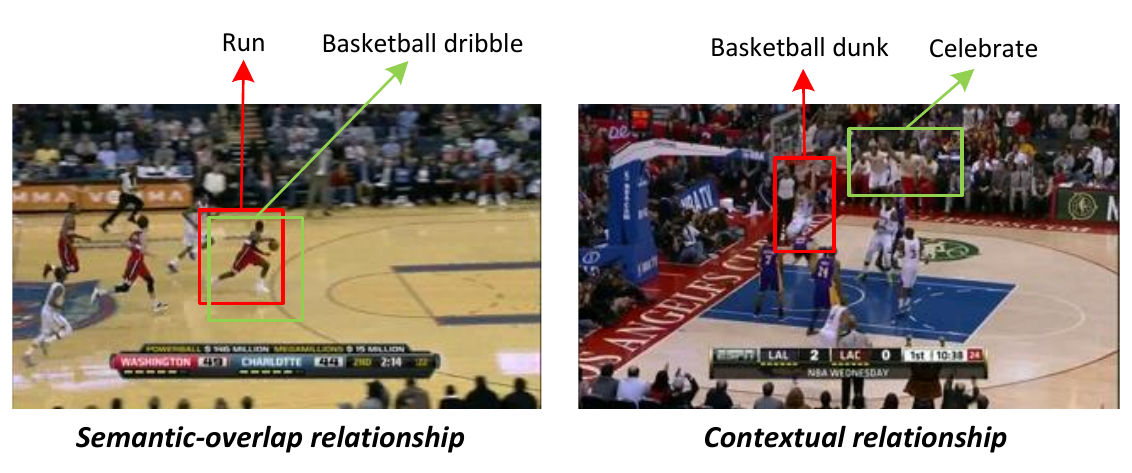}
	\caption{Examples of the co-occurrence relationship. The first example is the co-occurrence  of \textit{``Basketball dribble''} and \textit{``Run''}, which indicates the semantic overlap relationship. The second example is the co-occurrence of \textit{``Basketball dunk''} and \textit{``Celebrate''}, which indicates the contextual relationship.}
	\label{fig_1}
\end{figure}

Except for temporal modeling, the multi-label case, \textit{i.e.}, multiple actions occur at the same time, is also one of the most challenging problems for TAL, which often occurs in practice.
It is difficult for models to detect all these co-occurring actions.
There always exists some relationship among these co-occurring actions.
In general, as is shown in Figure \ref{fig_1}, the co-occurrence relationship among actions can be divided into two cases. One is the semantic overlap relationship and the other is contextual relationship.
For example, during a basketball game, the actions of \textit{``Basketball Dribble''} and \textit{``Run''} often share the same motion pattern.
And the actions of \textit{``Basketball Dunk''} and \textit{``Celebrate''} have close contextual relationship.
Understanding the relationship among these actions is significant to deal with the multi-label case.

There exist some works \cite{piergiovanni2018learning,tirupattur2021modeling} that focus on modeling the relationship among actions. 
Piergiovanii \textit{et al.} \cite{piergiovanni2018learning} propose a new concept \textit{``super event''} to relate a set of sub-events in terms of fixed temporal patterns to understand the relationship among actions.
Praveen \textit{et al.} \cite{tirupattur2021modeling} propose an attention-based multi-layer action dependency layer, which contains a co-occurrence dependency branch that models the co-occurrence relationship between different actions within a time step.
However, these methods all adopt an implicit modeling approach, \textit{i.e.}, the network learns the relationship among actions without any supervision. This causes two drawbacks. First, the network is likely to learn inadequate or even incorrect relationship. 
Second, it takes more time for the model to capture the correct relationship. 
Considering that the guidance of proper supervision is significant, we propose a method that can model the co-occurrence relationship among actions in an explicit manner. 
\IEEEpubidadjcol 

Besides, when modeling the relationship among actions, the existing works always ignore the semantic information among action classes and only focus on the visual information. 
Actually, the semantic relationship of co-occurrence actions is usually much closer.
For instance, action \textit{``Pole Vault''} is highly semantically related to action \textit{``Jump''}.
Also, action \textit{``Basketball Dribble''} and action \textit{``Running''} have a great semantic overlap.
On the other hand, semantic information has been proven to be helpful in other computer vision tasks, such as few-shot object detection \cite{zhu2021semantic,nie2022node,yang2022sega} and video gesture recognition \cite{wang2022beyond}.
However, it has not been explored in temporal action localization.
\begin{figure*}[!t]
	\centering
	\includegraphics[width=6in]{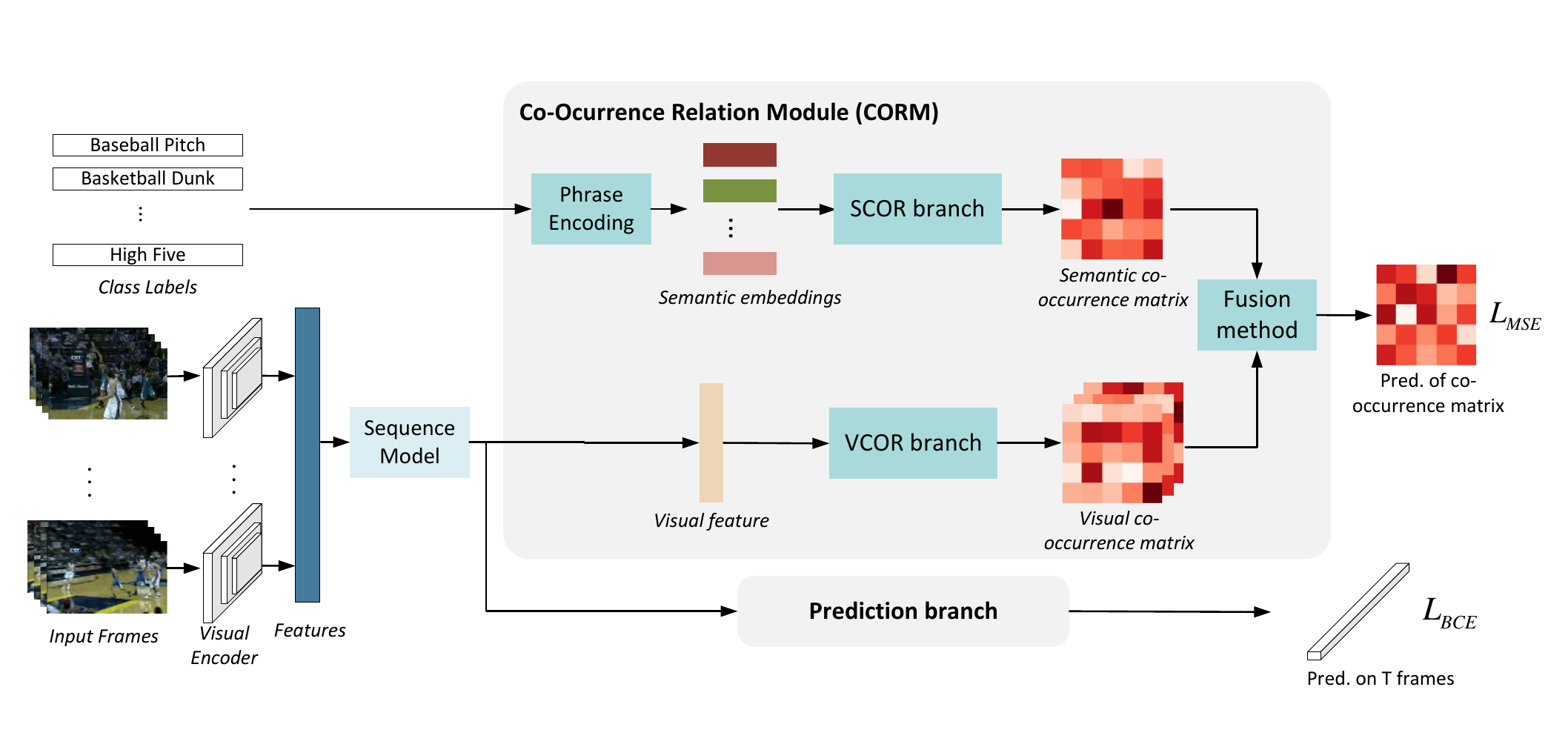}
	\caption{The structure of our COR Network. It contains a main prediction branch and a proposed Co-Ocurrence Relation Module (CORM). The input of CORM are class labels and the visual feature extracted from sequence model. The visual feature also acts as the input of the prediction branch. The CORM learns the co-occurrence relationship among actions and help the model to deal with the multi-label case. It plays a supporting role. And in inference, the TAL results can be obtained just according to the output of the prediction branch.}
	\label{cr_network}
\end{figure*}

Thus, in this work, we propose a novel lightweight Co-Occurrence Relation Module (CORM) to deal with the multi-label case, considering not only the visual but also the semantic information.
In the CORM, we propose to estimate the co-occurrence intensity of the actions relying both on the spatiotemporal visual features and on the semantic space of class labels.
The CORM contains two branches, \textit{i.e.} a Visual Co-Occurrence Relation branch (VCOR branch) and a Semantic Co-Occurrence Relation branch (SCOR branch).
The VCOR branch extracts the corresponding visual feature of each action, and then uses these features to capture the co-occurrence relationship. 
While the SCOR branch directly leverages the semantic embeddings of class labels to model the relationship.
To supervise the CORM, we construct a co-occurrence matrix as ground truth, which is calculated based on the class labels of each time without any extra annotations.
It reflects which actions frequently occur together in a video sequence.
With the supervision, the network can better understand the relationship among these co-occurrence actions.
CORM is plug-and-play.
As shown in Figure \ref{cr_network}, it can be combined with existing sequence models in TAL, \textit{i.e.,} models that only focus on temporal modeling, such as TGM \cite{piergiovanni2019temporal}, PDAN \cite{dai2021pdan}, to build Co-Occurrence Relation (COR) Networks.
Our proposed CORM enables COR Networks to integrate the co-occurrence relationship in the learned features of sequence models, to help the model make more accurate and comprehensive predictions.
It should be noted the CORM is only an auxiliary module which is used in training for better learning ability. 
And in inference, we only evaluate the performance of the prediction branch.

In addition, the current datasets \cite{yeung2018every,dai2022toyota} for TAL are always fine-grained and low-semantic. 
This makes the network lack the ability to understand complex events macroscopically. 
Therefore, we introduce a novel high-semantic dataset UCF-Crime-TAL based on UCF-Crime \cite{sultani2018real}, which is a dataset for anomaly detection containing 13 high-semantic anomaly classes. 
We annotate UCF-Crime at the frame level to form a multi-label video dataset UCF-Crime-TAL for temporal action localization.
When annotating, we label all classes that fit the real situation for each time step.
Although some works \cite{liu2019exploring,ozturk2021adnet} also label the UCF-Crime dataset along the temporal dimension, they don't consider the multi-label case and only assign one class label to a time step.
There are two main challenges within UCF-Crime-TAL: (1) It only focuses on abnormal events but not all subsistent events. 
(2) The anomaly classes within it are high-semantic.
For instance, the anomaly event \textit{``Arrest''} may be a series of element actions such as assault and shooting. This requires the model to abstract the concept of \textit{``Arrest''} from these continuous concrete actions. 
Thus this dataset is highly challenging. 

In summary, this paper makes the following contributions:
\begin{itemize}
	\item We propose a new Co-Occurrence Relation Module (CORM) that models the co-occurrence relationship among actions explicitly.
	\item We introduce the semantic information to the CORM and utilize it to model semantic relation among different actions based on the word embeddings of class labels.
	\item We introduce a new dataset UCF-Crime-TAL for the action localization task, which is annotated with dense labels considering the multi-label case. It is challenging due to its high-semantic action classes.
	\item We evaluate our method on three datasets, \textit{i.e.}, MultiTHUMOS, TSU, and our newly introduced UCF-Crime-TAL dataset. Our method consistently outperforms existing state-of-the-art methods.
\end{itemize}

\section{Related work}
\subsection{Temporal Action Localization}
Temporal action localization is popular due to its good fit to reality. There exist two main challenges in this field, \textit{i.e.}, the different ranges of actions' temporal durations and the multi-label case.
There mainly exist two kinds of methods in temporal action localization. 
The top-down methods \cite{DBLP:journals/tmm/ChenZZ21,DBLP:conf/iccv/XuDS17,lin2018bsn, liu2019multi, lin2019bmn,shou2017cdc,DBLP:conf/cvpr/HeilbronNG16,DBLP:conf/eccv/EscorciaHNG16} first propose temporal regions of actions, then identify their categories. Such methods require a large number of proposals and cause heavy computation costs.

The other bottom-up methods \cite{piergiovanni2018learning,tirupattur2021modeling,dai2021pdan,kahatapitiya2021coarse,piergiovanni2019temporal,dai2022ms,dai2019self,lea2017temporal} generate predictions for each time step and concatenate time steps with the same label to form the final proposals. These methods work faster. 
Existing bottom-up methods \cite{long2019gaussian,dai2021pdan,kahatapitiya2021coarse,dai2022ms} often focus on the first challenge.
There only exist a few works that focus on the second one.
Piergiovanni \textit{et al.} \cite{piergiovanni2018learning} define a concept \textit{``Super-event''} to connect a set of co-occurring sub-events with fixed temporal patterns.
Tirupattur \textit{et al.} \cite{tirupattur2021modeling} propose an M LAD layer that contains two sub-branches that model the actions' relationship within a timestep and the temporal pattern of each action respectively.
However, the above methods both model the action relation in an implicit manner, which means no supervision is provided. 
That leads to insufficient and inefficient learning ability.
Consequently, we propose the CORM which works in an explicit manner, \textit{i.e.}, it learns the co-occurrence relationship under the guidance of supervision, which is built based on the original temporal annotations of the datasets without any additional information.

It is important to note the difference between temporal action localization (TAL) and temporal action segmentation (TAS).
TAS is a very closely-related task that aims to segment a temporally untrimmed video and classify each segmented part.
It mainly deals with procedural sequences, such as \textit{``making coffee''}, \textit{``fried egg''} \textit{et al}.
While TAL focuses on more general videos where the temporal relation among actions is not such strong. 
And TAS focuses on finding the exact transition point between actions.
However, TAL pays more attention to the duration of each action.

\subsection{Semantic Information Model}
Obviously, the semantic information largely reflects the relationship between actions. 
However, in temporal action localization, it is ignored. 
While in the field of zero-shot object detection \cite{zhu2021semantic,nie2022node,yang2022sega,DBLP:conf/accv/ZhengHHHC20,DBLP:conf/cvpr/ZhuWS20}, the semantic relationship between objects is always utilized for auxiliary detection.
Zhu \textit{et al.} \cite{zhu2021semantic} propose to use the word embeddings of class labels to construct a semantic space to learn the relationship between different classes. Then it fuses this relationship into the network.
Nie \textit{et al.} \cite{nie2022node} learn the relationship among different classes from visual and semantic perspectives respectively. For the semantic relationship modeling, it constructs a semantic relational graph, where nodes are the semantic representations of classes.
Yang \textit{et al.} \cite{yang2022sega} use the semantic knowledge to enable the model to know what visual features should be focused on.
In gesture recognition, Wang \textit{et al.} \cite{wang2022beyond} attempt to fuse the semantic relationship into the network to assist gesture recognition.
However, this semantic information has not been considered in TAL.
As far as we know, we are the first to utilize the word embeddings of class labels to model semantic co-occurrence relationship in TAL.

\subsection{Datasets}
In recent years, researchers have proposed various datasets for temporal action localization.
At first, most datasets \cite{caba2015activitynet,THUMOS14,DBLP:journals/corr/LiuHLSL17,zhao2019hacs} focus on videos that contain sparse and well-separated instances of actions.
ActivityNet \cite{caba2015activitynet} and THUMOS \cite{THUMOS14} all have a large number of videos, which focus on sports and outdoor activities respectively.
It only exists a few action instances in a video. 
However, in real life, there always occur multiple actions at the same time. To better fit the reality, datasets with dense labels \cite{yeung2018every,dai2022toyota} are proposed.
MultiTHUMOS \cite{yeung2018every} is a dataset that extends THUMOS \cite{THUMOS14} from 20 action classes to 65 classes. 
And the label number is extended from 0.3 per frame to 1.5 per frame. 
TSU \cite{dai2022toyota} contains realistic untrimmed videos, which record the diverse spontaneous human activities in real-world settings.
The activities within the above datasets are all atomic and low-semantic, such as \textit{``Run''}, \textit{``Jump''}, \textit{etc}.
However, in real life, we always need to pay more attention to highly semantic activities. 
Motivated by the shortcomings of these datasets, we introduce UCF-Crime-TAL, which is formed by annotating the UCF-Crime \cite{sultani2018real} with dense labels at the frame level. 
When annotating, we consider the overlap of different events, thus making the multi-label annotations.
\section{Method}
In this section, we introduce our Co-Occurrence Relation Module (CORM). Firstly, we introduce how to build a co-occurrence matrix for a video sequence, which is used as the ground truth of the CORM. Then we describe the specific structure of the CORM.
Finally, we show the method to construct the COR Network and the design of the loss function.

\subsection{Ground truth building}
To model the co-occurrence relationship correctly and efficiently, we construct ground truth of this relationship for the CORM.
It should be noted that we only use the original temporal annotations of the datasets to build this ground truth, without any additional information.

We mathematically represent the co-occurrence relationship. Given a video sequence of length $T$, we build an $N\times N$ co-occurrence matrix $R^{*}$ as the ground truth of the CORM, where $N$ denotes the class number of actions. 
$R^{*}(i,j)$ represents the co-occurrence relation intensity between class $i$ and class $j$. 
The following is the detailed building process.
According to the original temporal annotations, we can get which class of action occurs at each time step.
Suppose that there occur $K$ classes of actions at time step $t$, we can construct an action class set $C_t$ based on their categories:
\begin{equation}
	C_t=\{c_k\}_{k=1}^{K}, c_k\in{1,2,\cdots,N}
\end{equation}
where $c_k$ denotes the class of the $k$-th action.
Then we build a co-occurrence matrix $R^{*}_{t}(i,j)$ for time step $t$:
\begin{equation}
	R^{*}_{t}(i,j)=\left\{
	\begin{aligned}
		1 ,&      & if \ i\in C_t \land  j\in C_t \\
		0 ,&      & otherwise
	\end{aligned}
	\right.
\end{equation}

In this manner, we can relate the co-occurring actions within each time step. Finally, to obtain the co-occurrence relationship for the video sequence, we sum up the co-occurrence matrixes along the temporal dimension:
\begin{equation}
	R^{*}=\sum_{t=0}^{T}R^{*}_{t}
\end{equation}
where $R^*$ is an $N \times N$ matrix that denotes the co-occurrence relationship within the whole video sequence. 

When two classes of actions occur simultaneously more often, the corresponding value in $R^{*}$ is higher, which indicates a closer co-occurrence relationship between them. With the supervision of this matrix, the network can better understand the relationship within the co-occurrence actions.

\subsection{Co-Occurernce Relation Module (CORM)}
\begin{figure}[!t]
	\centering
	\includegraphics[width=4in]{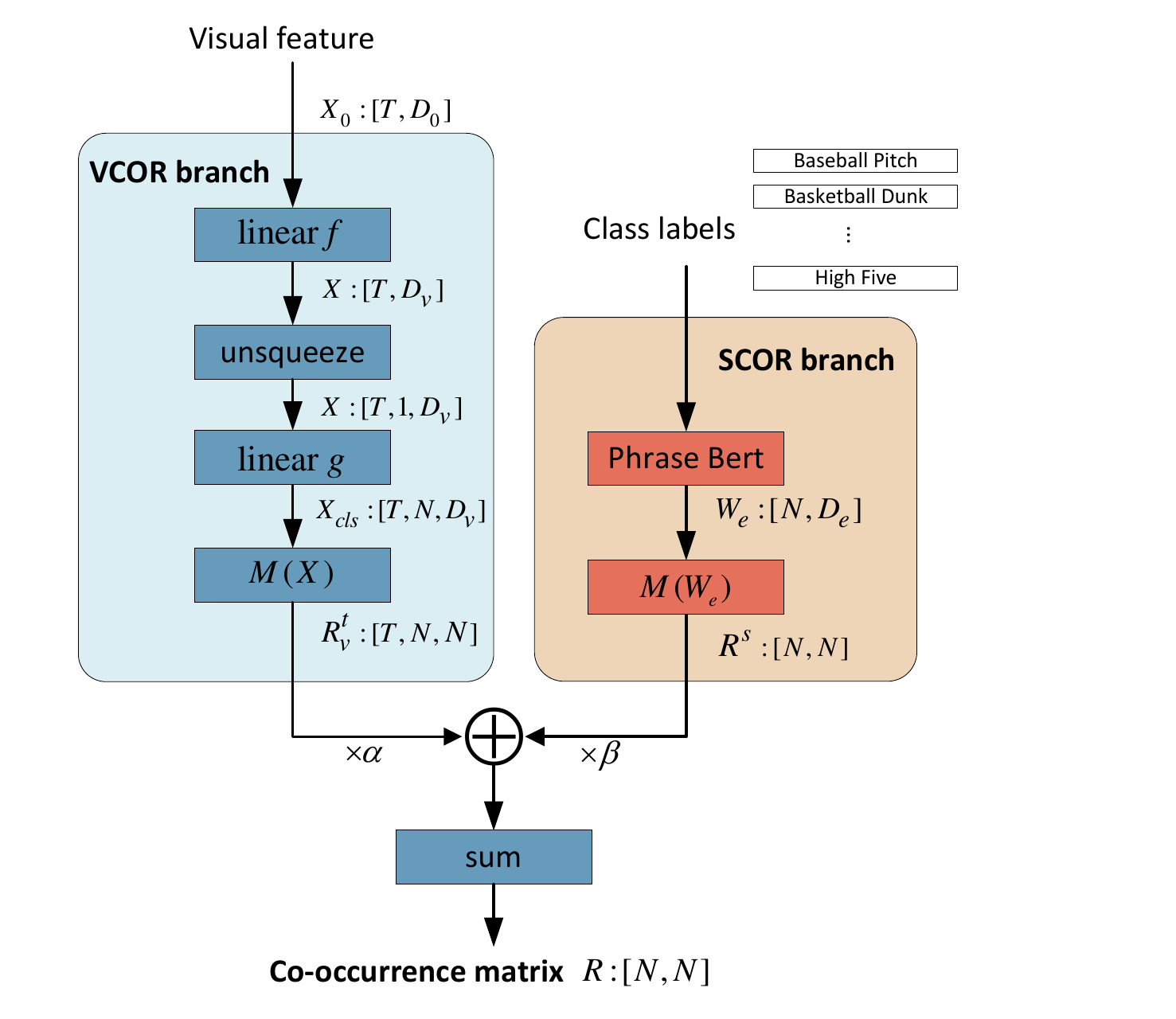}
	\caption{The structure of Co-Occurrence Relation Module (CORM). The CORM contains two branches, \textit{i.e.}, the VCOR branch and the SCOR branch. The input of VCOR branch is visual feature extracted from the sequence model. The input of SCOR branch is the class labels.}
	\label{vr_branch}
\end{figure}

CORM explores the co-occurrence relationship among actions to encode it in the learned features of the network. 
It contains two branches, \textit{i.e.}, the Visual Co-Occurrence Relation branch (VCOR branch) and the Semantic Co-Occurrence Relation branch (SCOR branch). 
The VCOR branch learns visual co-occurrence information. While the SCOR branch captures the co-occurrence relationship from the semantic perspective. 
The structure of the CORM is shown in Figure \ref{vr_branch}.

\subsubsection{Visual Co-Occurrence Relation branch}
The VCOR branch models the co-occurrence relationship among actions from the visual perspective.
Specifically, our VCOR branch contains three parts: (1) feature preprocessing, (2) class-specific feature extraction, and (3) correlation modeling.

In the first part, we preprocess the input feature $X_0\in R^{T\times D_0}$, where $D_0$ denotes the channel dimension. We utilize a linear layer $f$ to reduce the channel dimension from $D_0$ to $D_v$:
\begin{equation}
	X = f(X_0), X \in \mathbb{R}^{T\times D_v}
\end{equation}

Then, we unsqueeze $X$ to obtain the new feature $X'\in \mathbb{R}^{T\times 1\times D_v }$. 

Secondly, in the class-specific feature extraction part, given $X'$, we use a linear layer $g$ to extract corresponding features for each class:
\begin{equation}
	X_{cls} = g(X'), X_{cls}\in \mathbb{R}^{T\times N\times D_v}
\end{equation}
where $N$ denotes the number of classes. $X_{cls}(t,i,:)$ indicates the extracted feature for the $i$-th class in time step $t$. 

Finally, based on the class-specific features, we can obtain the co-occurrence matrix. For the class-specific feature $X_{cls}(t) \in \mathbb{R}^{N\times D_v}$ in time step $t$, we can compute their co-occurrence relationship by the following method:
\begin{equation}
	R^v_t = M(X_{cls}(t)), R^v_t\in \mathbb{R}^{N\times N} 
\end{equation}
where $R^v_t$ is the learned co-occurrence matrix at time step $t$. $M$ denotes the correlation modeling method, which will be discussed in Section \ref{section_m}.

\subsubsection{Semantic Co-Occurrence Relation branch}
The SCOR branch models the co-occurrence relationship among actions from the semantic perspective. 
In few-shot object detection, existing methods \cite{zhu2021semantic,nie2022node,yang2022sega} always utilize semantic information to model the relationship among different classes.
They always construct the semantic space using the Word2Vec \cite{DBLP:journals/corr/abs-1301-3781} or Glove \cite{pennington2014glove} which specifically encodes individual words. 
However, the class label in video tasks are always phrases, rather than individual words. 
Therefore, we utilize Phrase-BERT \cite{DBLP:conf/emnlp/WangTI21} that focuses on phrase embeddings to encode the action labels.
It proposes a contrastive finetuning objective to enable BERT \cite{DBLP:conf/naacl/DevlinCLT19} to encode more powerful phrase embeddings.

Given a set of class labels, we first encode them into embeddings using Phrase-BERT, thus constructing the semantic space $W_e \in R^{N\times D_e}$, where $W_e(i)$ is the semantic embedding of the $i$-th class label.  
For the semantic information, we also utilize the correlation modeling function $M$ to capture the co-occurrence relationship from the semantic perspective:
\begin{equation}
	R^s = M(W_e), R^s \in \mathbb{R} ^{N\times N}
\end{equation}

This branch delves into the semantic relationship among actions. The correlation modeling function will be discussed in Section \ref{section_m}

\subsubsection{Fusion method}
After capturing the co-occurrence relationship from visual and semantic perspectives, we fuse them to obtain a combined co-occurrence relationship.
We firstly add the semantic co-occurrence matrix with the visual co-occurrence matrix in each time step by a weighted summation.
\begin{equation}
	R_t = \alpha * R^v_t+ \beta * R^s
	\label{weighted_sum}
\end{equation}
where the $\alpha$ and $\beta$ are both learnable parameters, which indicate the importance of $R^v$ and $R^s$ respectively.
Then we sum the $R_t$ along the temporal dimension to obtain the co-occurrence matrix for the whole video sequence:
\begin{equation}
	R=\sum_{t=0}^{T}R_{t}, R \in \mathbb{R}^{N \times N}
\end{equation}

By the fusion method discussed above, we can obtain the co-occurrence matrix which contains both visual and semantic information.

\begin{figure}[!t]
	\centering
	\subfloat[$M_1(X)$]{
		\begin{minipage}[t]{\linewidth}
			\centering
			\includegraphics[width=3.2in]{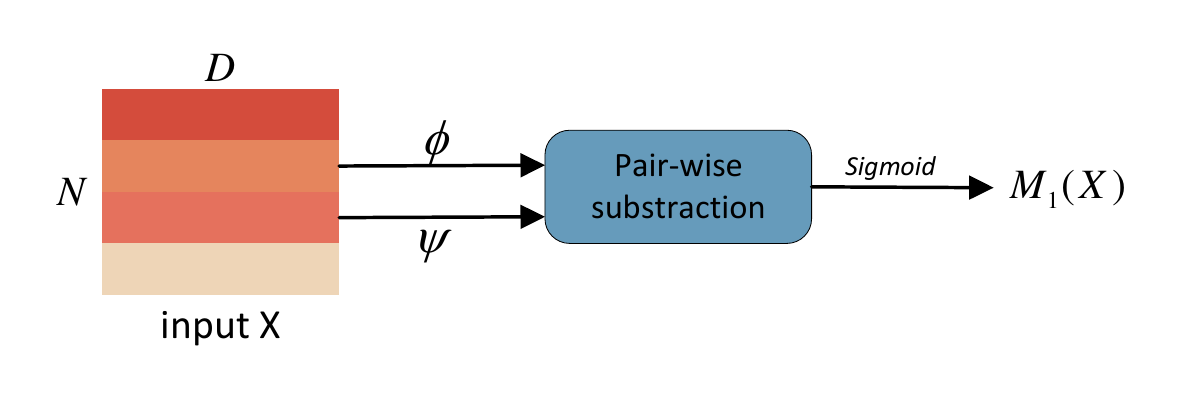}
		\end{minipage}
	}
	
	\subfloat[$M_2(X)$]{
		\begin{minipage}[t]{\linewidth}
			\centering
			\includegraphics[width=3.2in]{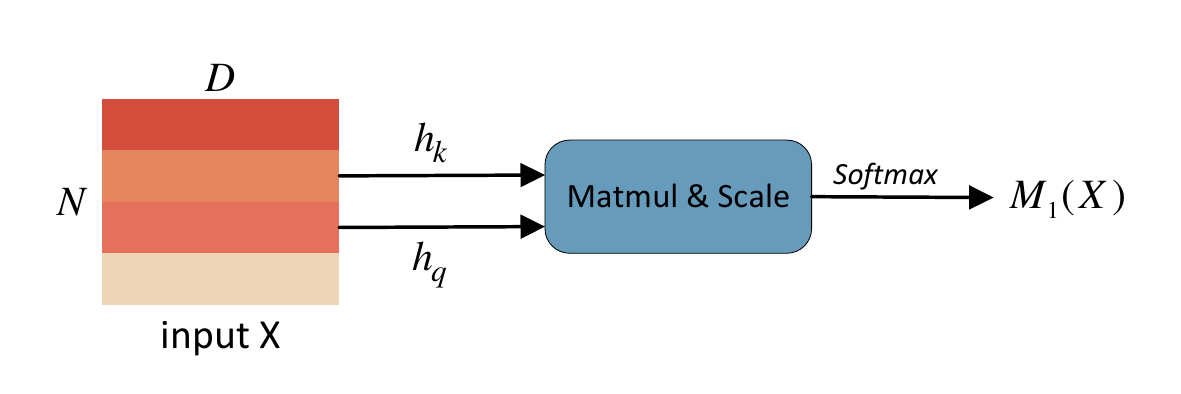}
		\end{minipage}
	}%
	\centering
	\caption{Two functions to model the co-occurrence relationship. Figure (a) is the first function and Figure (b) is the second one. }
	\label{m_function}
\end{figure}

\subsubsection{Correlation Modeling function} \label{section_m}

In this section, we discuss the correlation modeling function $M$. We propose two methods, which are shown in Figure \ref{m_function}.

The first method maps the input feature $X\in \mathbb{R}^{N\times D}$ to a new space and then computes the distance among different actions as their correlation strength. Specifically, given a pair of classes ($i$, $j$) and their corresponding features ($x_i$, $x_j$), we formulate the first correlation modeling function $M_1$ as:
\begin{equation}
	M_1(x_i,x_j) = \sigma (\phi(x_i)-\psi(x_j))
\end{equation}
where $\sigma(\cdot)$ indicates the sigmoid activation function.
$\phi(\cdot)$ and $\psi(\cdot)$ are linear layers which reduce the channel dimension from $D$ to 1.
$M_1(\cdot)$ calculates the distance between $\phi (x_i)$ and $\psi(x_j)$, which indicates the co-occurrence relationship between the $i$-th class and the $j$-th class.

The second method is based on the self-attention mechanism. Given an input $X\in R^{N\times D}$, we first transform it into $Q$ and $K$ via two linear layers $h_q$ and $h_k$. Then, we calculate the self-attention matrix using $Q$ and $K$:
\begin{equation}
	h_q, h_k : \mathbb{R}^{N\times D} \mapsto \mathbb{R}^{N \times d_{k}}, Q=h_q\cdot X, K=h_k \cdot X
\end{equation}

\begin{equation}
	M_2(X)=softmax(\frac{QK^{T}}{\sqrt{d_k}})
\end{equation}
where the element in $(i,j)$ of $M_2(X)$ represents the co-occurrence relationship between class $i$ and class $j$.

\subsection{COR Network}
\begin{figure*}[!t]
	\centering
	\subfloat[]{\includegraphics[width=3.2in]{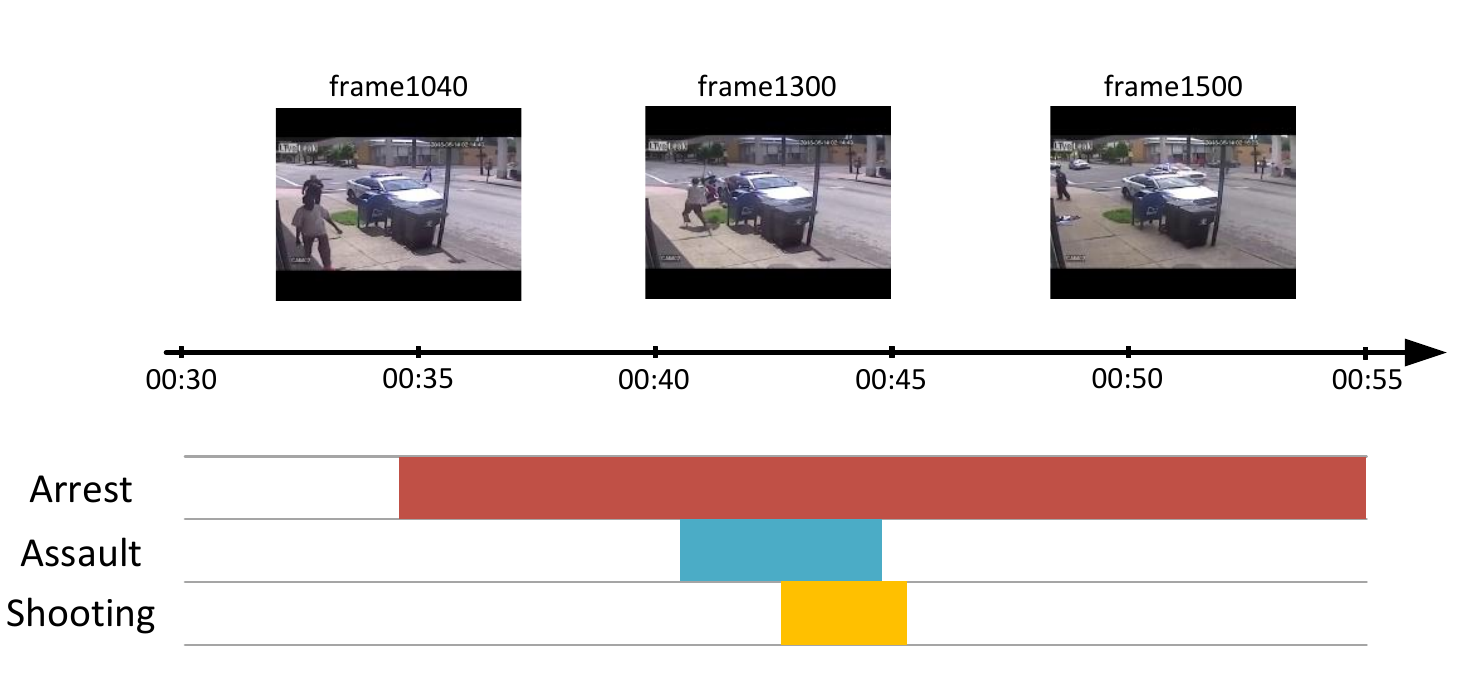}%
		\hspace{0.7cm}
		\label{dataset_example}}
	\subfloat[]{\includegraphics[width=3.4in]{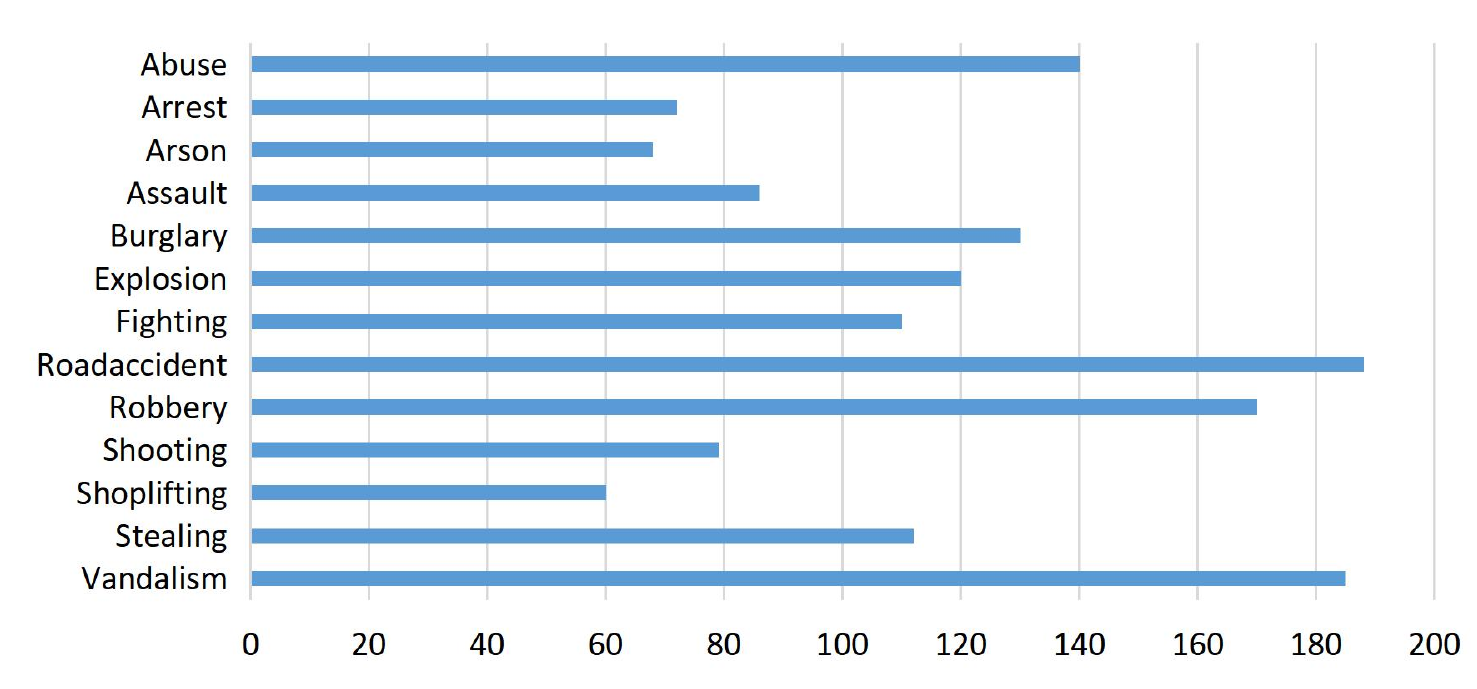}%
		\label{distribution}}
	\caption{(a) An example of annotations on UCF-Crime-TAL dataset. It shows a case of \textit{``Arrest''} in which the action of \textit{``Assault''} and \textit{``Shooting''} also take place. (b) The distribution of different classes in UCF-Crime-TAL dataset.}
\end{figure*}
We insert the CORM to sequence models to build our Co-Occurrence Relation Network (COR Network) for temporal action localization. In previous works, after obtaining the video sequence feature, the networks always only use the feature to make predictions for each time step. Different from them, we also utilize it to capture the co-occurrence relationship among actions, \textit{i.e.}, this feature is also fed into our CORM.
The CORM is used in training to help the model capture the co-occurrence relationship among actions. 
And when inference, the TAL results can be obtained from the prediction branch. 
Thus during inference, we just need to evaluate the performance of the prediction branch.

As shown in Figure \ref{cr_network}, our COR Network contains a
main prediction branch and a auxiliary module CORM.
The extracted feature $X_0\in R^{T\times D_0}$ from the sequence model is fed into the prediction branch as well as the CORM.
The structure and ground truth of CORM are already discussed above. 
The prediction branch utilizes a linear layer to predict class labels for each time step. Specifically, given the extracted feature of length $T$, each time-step $t$ contains a ground-truth label $y_{t,c}\in \{0,1\}$, where $c\in {1,\dots,N}$ indicates the action class. The prediction branch needs to predict class probabilities $\tilde{y}_{t,c}\in [0,1]$.

We design loss functions for our COR Network. For the prediction branch, we use the Binary Cross Entropy (BCE) loss. For the CORM, we utilize the Mean Squared Error (MSE) loss to supervise. Specifically, the two loss functions are described as follows:
\begin{equation}
	L_{BCE} = y_{t,c}\log(p(c|v_t))+(1-y_{t,c})\log(1-p(c|v_t))
\end{equation}

\begin{equation}
	L_{MSE} = \frac{1}{n}\sum_{i=1}^{N}(R^*_i-R_i)^2
\end{equation}
where $v_t$ is the visual feature at frame $t$ and $y_{t,c}$ is the ground-truth label for class $c$ in time $t$. The $R_i$ and $R^*_i$ indicate the prediction and the ground truth of the co-occurrence relation intensity between action $i$ and other actions. 

The overall loss is computed by a weighted sum of the two losses introduced above.
\begin{equation}
	L_{total} =  L_{BCE} + a\times L_{MSE}
\end{equation}
where $a$ is used to control the numerical scale of the MSE loss.

\section{UCF-Crime-TAL Dataset}
In this section, we introduce our proposed UCF-Crime-TAL dataset.
We first describe how we densely annotate the UCF-Crime \cite{sultani2018real} dataset at the frame level.
Then we discuss the challenges of the UCF-Crime-TAL dataset.

\subsection{Annotation}

The UCF-Crime \cite{sultani2018real} dataset is a large-scale video anomaly detection dataset, which contains 13 anomaly classes.
This dataset consists of not only abnormal videos but also normal videos where no abnormal events occur.
UCF-Crime only has video-level abnormal-or-not labels, \textit{i.e.,} whether the current video has abnormal events or not, without the specific class of the abnormal events.
When we use UCF-Crime for TAL, we only care about abnormal events.
The normal events act as the background.
Therefore, to better match the temporal action localization task, we remove videos which only contain normal events to construct our UCF-Crime-TAL dataset.
Then we annotate the UCF-Crime-TAL dataset with dense labels at the frame level.
Given a video, we first locate the abnormal events. 
Then we annotate each time step with all abnormal classes that occur currently.
Figure \ref{dataset_example} shows an example of the annotation. 
This video shows an example of action \textit{``Arrest''}. When the policeman arrests the criminal, the criminal assaults the policeman. Then the policeman shoots at him. 
And we show the distribution of abnormal activities in UCF-Crime-TAL in Figure \ref{distribution}.

\subsection{Challenges}
The UCF-Crime-TAL dataset is based on the UCF-Crime \cite{sultani2018real} dataset, which aims at anomaly detection. 
This leads to 2 challenges.

(1) \textbf{Only focus on abnormal events:} It only focuses on abnormal events rather than all occurred events. 
In TAL, we need to localize the interested actions and classify them. 
Different from previous datasets, the UCF-Crime-TAL dataset only concerns about the anomaly events other than all subsistent events.
Thus to make correct predictions, networks need to distinguish which type of events should be focused on.

(2) \textbf{High-semantic level:} Compared to other datasets, UCF-Crime-TAL is more high-semantic. 
The abnormal events in it usually contain a series of atomic events. When making predictions, networks need to associate these atomic events to abstract out higher semantic-level abnormal events.
For example, as is shown in Figure \ref{dataset_example}, the criminal notices that the policeman is approaching him. So the criminal assaults the policeman. Then the policeman shoots at the criminal. In this case, the above atomic events form the high-semantic abnormal event \textit{``Arrest''}.
Only understand the semantic relationship among these atomic events can the model understand the high-semantic abnormal events.

\section{Experiments}
\subsection{Datasets}
We evaluate our method on two commonly used densely-labeled datasets MultiTHUMOS \cite{yeung2018every}, TSU \cite{dai2022toyota} and our proposed UCF-Crime-TAL dataset.
MultiTHOMOS is extended from THUMOS \cite{THUMOS14}, which contains videos of various sports activities. It is labeled with 65 different classes at frame-level.
It contains 413 videos where 200 of that are used for training and others for validation.
On average, MultiTHUMOS contains 1.5 labels per frame and 10.5 action classes per video.
TSU \cite{dai2022toyota} focuses on long untrimmed videos, the average length of which is 21 minutes. It contains 536 videos with 51 classes. We use its defined \textit{Cross-Subject} evaluation protocol for our work.
Specifically, it splits 18 subjects into training set and testing set, which contain 11 and 7 subjects respectively.
As for the UCF-Crime-TAL dataset, we follow the split setting of the training set and the testing set in UCF-Crime.

\subsection{Implementation Details}
Following the previous works, we utilize the I3D features as input.
They are the output features after the Global Average Pooling of the I3D \cite{carreira2017quo} network. 
Thus the dimension of input features is 1024.
For MultiTHUMOS and UCF-Crime-TAL datasets, the length of input features is fixed to 256.
As for TSU, because of its long duration, the length of input features is fixed to 512.
In our work, we insert the CORM into other sequence models to form different COR Networks.
When training our COR Network, we set the initial learning rate to 0.0005.
The batch size is set to 16, 8, and 32 for MultiTHUMOS, TSU, and UCF-Crime-TAL respectively. 
And we train 200, 300, and 60 epochs for them respectively.
Meanwhile, the loss balance factor $a$ is set to 0.001, 0.0003, and 0.0005 for MultiTHUMOS, TSU and UCF-Crime-TAL dataset respectively. 
When testing, we evaluate the per-frame mAP of these datasets.
\begin{table*}[!t]
	\caption{Comparison with the state-of-the-art methods on three datasets. We report the per-frame mAP only using RGB videos as input.}
	\centering
	\begin{tabular}{c c c c c c}
		\toprule
		Methods& Visual Encoder &GFLOPs& MultiTHUMOS & TSU & UCF-Crime-TAL \\
		\midrule
		I3D + LSTM \cite{DBLP:conf/cvpr/MahasseniT16} & I3D&--&29.9&15.9&-- \\
		Super-event \cite{piergiovanni2018learning} & I3D&0.8 &36.4&17.2& --\\
		TGM \cite{piergiovanni2019temporal} & I3D &1.2& 37.2& 26.7& --\\
		PDAN \cite{dai2021pdan} & I3D &3.2&40.2& 32.7&15.88 \\
		MLAD \cite{tirupattur2021modeling} & I3D&44.8 &42.2& --&15.18\\
		MS-TCT \cite{dai2022ms} & I3D&6.6&43.1&33.7&18.54\\
		\midrule
		baseline & I3D&5.76& 42.32& 30.69 & 18.07\\
		COR Network & I3D&5.79 &\bf{45.33}& \bf{36.34}&\bf{20.40}\\
		\bottomrule
	\end{tabular}
	
	\label{sota}
\end{table*}

\begin{table}[!t]
	\caption{The correlation modeling function choices of CORM. We choose different correlation modeling functions for VCOR branch and SCOR branch. And we evaluate the mAP value of these models. }
	\centering
	\begin{tabular}{c|c c| c c| c}
		\toprule
		Methods&\multicolumn{2}{c|}{VCOR branch}& \multicolumn{2}{c|}{SCOR branch}& \multirow{2}{*}{mAP} \\
		\cline{1-5}
		COR Network& $M1$ & $M2$ & $M1$ & $M2$&  \\
		\midrule
		A & \checkmark & & \checkmark & & 44.77 \\
		B & \checkmark & &  & \checkmark& 45.33 \\
		C &  & \checkmark& \checkmark & & 42.35 \\
		D &  & \checkmark&  & \checkmark& 43.24 \\
		\midrule
		baseline &--&--&--&--& 42.32 \\
		\bottomrule
	\end{tabular}
	\label{correlation_function}
\end{table}

\subsection{Comparison with the State-of-the-Art}

In this section, we compare our proposed COR Network with state-of-the-art TAL methods. 
The results are shown in Table \ref{sota}.
We use MS-TCT without the hmap branch as the baseline, in which we use 3 Global-Local Relation Blocks for each stage.
And we build our COR Network based on it.
For a comprehensive comparison, we reproduce several previous works on our UCF-Crime-TAL dataset.
Obviously, our COR Network achieves state-of-the-art performance on the three datasets.
Previous works \cite{DBLP:conf/cvpr/MahasseniT16,piergiovanni2018learning,piergiovanni2019temporal,dai2021pdan,dai2022ms,tirupattur2021modeling} always pay more attention to temporal modeling but not the multi-label case.
Superevent \cite{piergiovanni2018learning} and TGM \cite{piergiovanni2018learning} all propose new filters to model the temporal patterns of different actions.
PDAN \cite{dai2021pdan} and MS-TCT \cite{dai2022ms} all use multi-stage architecture to process different ranges of actions.
And MS-TCT fuses features from different stages to make predictions, thus it can achieve better performance.
Thanks to the above works, the problem of temporal modeling has been well solved.
However, the multi-label case is still a challenge.
MLAD \cite{tirupattur2021modeling} firstly focuses on the multi-label case, which utilizes multiple self-attention layers to model co-occurrence dependencies and temporal action dependencies simultaneously.
However, it learns these dependencies in an implicit manner, \textit{i.e.,} without any supervision.
That makes the model takes more time to capture the right dependencies.
Specifically, MLAD needs to train 2500 epochs to achieve its best performance.
Meanwhile, since it stacks multiple self-attention layers, it evolves much more computation costs (44.8 GFLOPs).
Different from it, we propose a lightweight plug-and-play module CORM, which can be easily incorporated with existing sequence models to handle both temporal modeling and the multi-label case.
Therefore, our CORM can achieve the best performance.
Compared with the previous works, our COR Network outperforms them by a large margin, \textit{i.e.}, 2.23 mAP on MultiTHUMOS, 2.64 mAP on TSU and 1.86 mAP on UCF-Cime-TAL respectively.

On the other hand, when inserting CORM into the baseline, the performance achieves a large amount of improvement, \textit{i.e.}, 3.01 mAP, 5.65 mAP, and 2.33 mAP on the three datasets respectively.
The baseline only focuses on the temporal modeling but does not take the co-occurrence relationship into account.
Obviously, CORM makes up for the shortcomings of the baseline.

Last but not least, our CORM is a lightweight module.
CORM only introduces a small number of parameters (0.05M) and GFLOPs (0.03).
The number of parameters of the baseline, MS-TCT, and COR Network are 83.93M, 87.11M, and 83.98M respectively. 
As shown in Table \ref{sota}, our COR Network achieves better performance with lower GFLOPs (5.79) and fewer parameters (83.98M) compared with MS-TCT (6.6, 87.11M). 
It also can be combined with other more efficient sequence models, which will be discussed in Section \ref{universality}.



\subsection{Ablation Study}

In this section, we perform full ablation study on the MultiTHUMOS dataset.
Firstly, we investigate the best setting of our CORM in Section \ref{correlation_setting}. Then, we study the effectiveness of each component in our CORM in Section \ref{component}.
Finally, we show the universality of CORM in Section \ref{universality}.
Note that \textit{baseline} in this section indicates MS-TCT without its hmap branch. And the COR Network in Section \ref{component} and Section \ref{universality} follows the best settings discussed in Section \ref{correlation_setting}.

\begin{table}[t]
	\caption{The effectiveness of each component in CORM. We add the VCOR branch and the SCOR branch to our baseline one by one and evaluate their mAP value on the MultiTHUMOS dataset. Baseline* indicates only adding the VCOR branch to the baseline.}
	\centering
	\begin{tabular}{c|c |c| c}
		\toprule
		Methods & VCOR branch& SCOR branch & mAP \\
		\midrule
		baseline & & & 42.32 \\
		baseline* & \checkmark & & 44.31 \\
		COR Network &   \checkmark& \checkmark & 45.33 \\
		\bottomrule
	\end{tabular}
	\label{effective}	
\end{table}

\subsubsection{Correlation methods}
\label{correlation_setting}
Based on the baseline, we apply different correlation methods to VCOR branch and SCOR branch to construct four different COR Networks. The results are shown in Table \ref{correlation_function}.
Compared with the baseline, except for COR Network C, other COR Networks all achieve obvious improvements, \textit{i.e.}, 2.45, 3.01 and 0.92 mAP respectively.
That indicates the effectiveness of modeling the co-ocurrence relationship, regardless of the specific approach. 
Among them, COR Network B achieves the best performance (45.33 mAP) which uses $M_1$ correlation function for the VCOR branch and $M_2$ for the SCOR branch.
Compared COR Network A(B) with C(D), we can find that the $M_1$ method performs better in the VCOR branch. 
While compared COR Network A(C) with B(D), obviously, the $M_2$ method performs better in the SCOR branch.
We suspect the reason is that $M_2$ correlation modeling method utilizes the self-attention mechanism, which is originally proposed in the field of natural language processing.
Thus, it may be more suitable to handle the input of word embeddings.


\subsubsection{Analysis on each component}
\label{component}
We analyze the effectiveness of each component in our CORM.
Specifically, we add the VCOR branch and SCOR branch one by one to the baseline to form three models. The results are shown in Table \ref{effective}.
And \textit{baseline*} indicates only adding the VCOR branch to the baseline.
We find that adding the VCOR branch and the SCOR branch can respectively bring 2 mAP and 1 mAP improvement to the model.
That indicates the effectiveness of our VCOR branch and SCOR branch.
When learning the co-occurrence relationship, the visual information plays a leading role.
And the semantic relationship among actions is an important auxiliary component which contains prior information.
The SCOR branch introduces a priori information of external knowledge, which can offer the semantic relationship that is missing in visual information.
Then during the training process, the SCOR branch can gradually fuse the hidden semantic relationship into CORM.
Note that it is unnecessary to evaluate the CORM that only contains the SCOR branch. 
Because the SCOR branch is directly based on the class labels other than the visual information. 
If we only use the SCOR branch in CORM, there will be no information backpropogated into the sequence model and no influence on the prediction branch. 
The prediction result is the same as the baseline. 
\begin{table}[t]
	\caption{Different sequence models w/ and w/o the CORM. We report the mAP value of these models on MultiTHUMOS dataset.}
	\centering
	\begin{tabular}{c c c}
		\toprule
		Methods & mAP & $\Delta$ mAP \\
		\midrule
		TGM & 37.2&\multirow{2}{*}{1.37} \\
		TGM + CORM & 38.57& \\
		\midrule
		PDAN &40.2& \multirow{2}{*}{1.33} \\
		PDAN + CORM &41.53& \\
		\midrule
		baseline &42.32& \multirow{2}{*}{3.01}\\
		baseline + CORM &45.33& \\
		\bottomrule
	\end{tabular}
	\label{general}
\end{table}

\subsubsection{Generalization of CORM}
\label{universality}
To demonstrate the generalization of the CORM, we insert it into 3 existing sequence models, \textit{i.e.}, TGM \cite{piergiovanni2019temporal}, PDAN \cite{dai2021pdan} and MS-TCT \cite{dai2022ms} without the hmap branch.
Then we report the mAP values of these models, which is shown in Table \ref{general}.
After inserting the CORM, the performance of the three sequence models is improved by 1.37, 1.33, and 3.01 mAP respectively.
As discussed before, the three sequence models all only focus on temporal modeling but neglect the co-occurrence relationship among actions.
When adding the CORM, the networks can handle both the temporal modeling problem and the multi-label case.
Thus they can get better performance.
And the improvement of these sequence models demonstrates that temporal modeling and co-occurrence relationship modeling are complementary, which are both significant for temporal action localization.
Last but not least, the improvement indicates that our CORM can be easily combined with existing sequence models and make up for their shortcomings in co-occurrence modeling.
\begin{table}[!t]
	\caption{The learned weight values in fusion method ($\alpha$ and $\beta$ in Equation \ref{weighted_sum})	on three datasets, \textit{i.e.,} MultiTHUMOS, TSU and UCF-Crime-TAL.}
	\centering
	\begin{tabular}{c c c}
		\toprule
		Datasets & $\alpha$ & $\beta$ \\
		\midrule
		MultiTHUMOS &0.0964&0.0463\\
		TSU &0.2095  &0.097 \\
		UCF-Crime-TAL &0.9544&0.9546 \\
		\bottomrule
	\end{tabular}
	\label{parameter}
\end{table}
\subsection{The values of $\alpha$ and $\beta$}
In this subsection, we discuss the values of the learned parameters, \textit{i.e.,} $\alpha$ and $\beta$ in the CORM.
We show the learned values of $\alpha$ and $\beta$ in Table \ref{parameter}.
We can find that in MulthiTHUMOS and TSU datasets, the values of $\alpha$ are both larger than the values of $\beta$.
That indicates the leading role of visual information, and the semantic information acts as a complementary role in these two datasets.
However, in our proposed UCF-Crime-TAL dataset, the values of $\alpha$ and $\beta$ are about the same.
We suspect it is due to the high-semantic level of class labels in UCF-Crime-TAL.
We believe that the values of $\alpha$ and $\beta$ imply the importance of the visual and semantic information.
When the class labels are high-semantic, it is much more difficult to model the co-occurrence relationship among actions only depending on the visual information.
And the semantic information contains prior information, which can guide the co-occurrence relationship modeling to a certain extent.
Thus at this point, the semantic information is more important, even on par with visual information.
In general, the suitable values of $\alpha$ and $\beta$ can help the model better integrate the visual and semantic co-occurrence information into the CORM, so that the model can learn more comprehensive co-occurrence relationship.
\begin{figure*}[!t]
	\centering
	\includegraphics[width=7in]{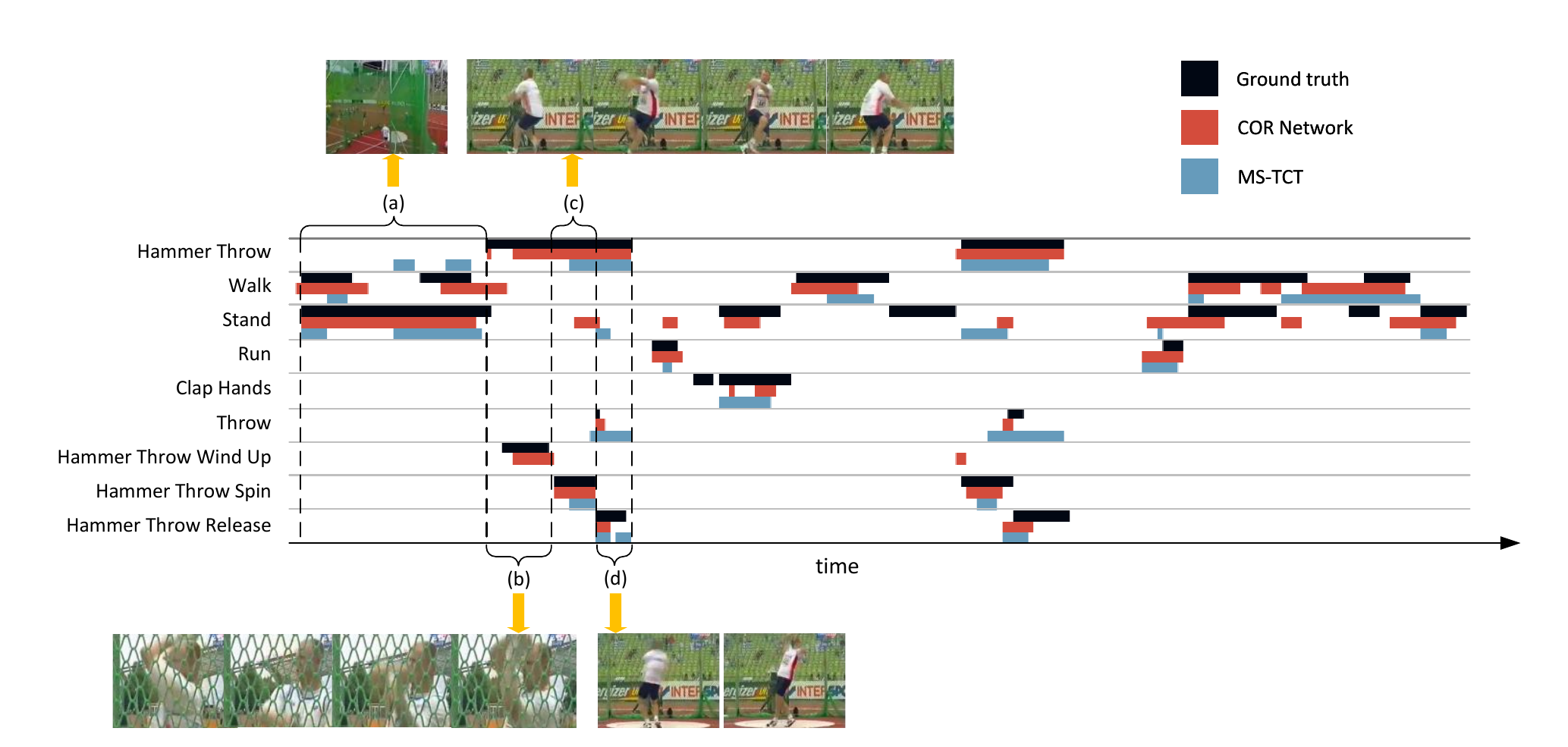}
	\caption{Visualization of the localization result on an example video from MultiTHUMOS dataset.
		In this figure, we visualize the ground truth and the localization results of MS-TCT and our COR Network.}
	\label{visual_detection}
\end{figure*}

\subsection{Visualization}
We visualize the localization result of an example on the MultiTHUMOS \cite{yeung2018every} dataset in Figure \ref{visual_detection}.
Compared with the MS-TCT, the localization result of our COR Network is much closer to the ground truth.
In particular, for some actions with shorter duration, our COR Network performs better than the MS-TCT.
And its prediction of the actions' durations is more consistent with the ground truth.
And most importantly, as is shown in the duration (a), (b), (c), and (d) in Figure \ref{visual_detection}, the COR Network performs better in the multi-label case.
In the duration (a), the hammer thrower is walking to the court.
And the class labels are \textit{``Walk''} and \textit{``Stand''}.
The semantic overlap between \textit{``Walk''} and \textit{``Stand''} is high.
Compared with MS-TCT, our COR Network makes more accurate predictions.
It indicates our COR Network can better dig out the semantic-overlap relationship among the co-occurring actions. 
And for the durations (b), (c), and (d), they are actually three subactions of the action \textit{``Hammer Throw''}.
The hammer thrower first winds up the hammer.
Then he spins his body to accelerate the hammer.
Finally, he releases the hammer at the right time.
The contextual relationship among \textit{``Hammer Throw''}, \textit{``Hammer Throw Wind Up''}, \textit{``Hammer Throw Spin''}, and \textit{``Hammer Throw Release''} is very tight.
Our COR Network explicitly models the co-occurrence relationship among actions, so it performs better in this case.

\section{Conclusion}
In this work, we propose a novel module CORM to solve the multi-label case in TAL by learning the co-occurrence relationship among actions explicitly with the guiance of the actions' co-occurrence matrix.
It works in a plug-and-play manner and can be easily incorporated with the existing sequence models.
It contains a VCOR branch and a SCOR branch, which model the co-occurrence relationship among actions both from the visual perspective and the semantic perspective.
Considering the existing datasets in TAL are low-semantic, we construct a new high-semantic dataset UCF-Crime-TAL by annotating the UCF-Crime dataset for every time step with dense labels.
Finally, we evaluate our proposed CORM on three challenging multi-label TAL datasets, on which it achieves state-of-the-art performance consistently.

%
%
%


%
 \bibliography{reference.bib}

\begin{thebibliography}{10}
\providecommand{\url}[1]{#1}
\csname url@samestyle\endcsname
\providecommand{\newblock}{\relax}
\providecommand{\bibinfo}[2]{#2}
\providecommand{\BIBentrySTDinterwordspacing}{\spaceskip=0pt\relax}
\providecommand{\BIBentryALTinterwordstretchfactor}{4}
\providecommand{\BIBentryALTinterwordspacing}{\spaceskip=\fontdimen2\font plus
\BIBentryALTinterwordstretchfactor\fontdimen3\font minus
  \fontdimen4\font\relax}
\providecommand{\BIBforeignlanguage}[2]{{%
\expandafter\ifx\csname l@#1\endcsname\relax
\typeout{** WARNING: IEEEtran.bst: No hyphenation pattern has been}%
\typeout{** loaded for the language `#1'. Using the pattern for}%
\typeout{** the default language instead.}%
\else
\language=\csname l@#1\endcsname
\fi
#2}}
\providecommand{\BIBdecl}{\relax}
\BIBdecl

\bibitem{lin2018bsn}
T.~Lin, X.~Zhao, H.~Su, C.~Wang, and M.~Yang, ``Bsn: Boundary sensitive network
  for temporal action proposal generation,'' in \emph{Proceedings of the
  European Conference on Computer Vision (ECCV)}, 2018, pp. 3--19.

\bibitem{liu2019multi}
Y.~Liu, L.~Ma, Y.~Zhang, W.~Liu, and S.-F. Chang, ``Multi-granularity generator
  for temporal action proposal,'' in \emph{Proceedings of the IEEE/CVF
  Conference on Computer Vision and Pattern Recognition}, 2019, pp. 3604--3613.

\bibitem{tirupattur2021modeling}
P.~Tirupattur, K.~Duarte, Y.~S. Rawat, and M.~Shah, ``Modeling multi-label
  action dependencies for temporal action localization,'' in \emph{Proceedings
  of the IEEE/CVF Conference on Computer Vision and Pattern Recognition}, 2021,
  pp. 1460--1470.

\bibitem{9660459}
J.~Zhang and J.~Yin, ``Enhancing class-semantics features' locating performance
  for temporal action localization,'' in \emph{2021 7th IEEE International
  Conference on Network Intelligence and Digital Content (IC-NIDC)}, 2021, pp.
  259--263.

\bibitem{9454500}
B.~Wang, X.~Zhang, and Y.~Zhao, ``Exploring sub-action granularity for weakly
  supervised temporal action localization,'' \emph{IEEE Transactions on
  Circuits and Systems for Video Technology}, vol.~32, no.~4, pp. 2186--2198,
  2022.

\bibitem{8440749}
H.~Song, X.~Wu, B.~Zhu, Y.~Wu, M.~Chen, and Y.~Jia, ``Temporal action
  localization in untrimmed videos using action pattern trees,'' \emph{IEEE
  Transactions on Multimedia}, vol.~21, no.~3, pp. 717--730, 2019.

\bibitem{DBLP:conf/ijcai/YangW22}
S.~Yang and X.~Wu, ``Entity-aware and motion-aware transformers for
  language-driven action localization,'' in \emph{Proceedings of the
  Thirty-First International Joint Conference on Artificial Intelligence},
  L.~D. Raedt, Ed., 2022, pp. 1552--1558.

\bibitem{DBLP:journals/tcsv/SunSYX23}
W.~Sun, R.~Su, Q.~Yu, and D.~Xu, ``Slow motion matters: {A} slow motion
  enhanced network for weakly supervised temporal action localization,''
  \emph{IEEE Transactions on Circuits and Systems for Video Technology},
  vol.~33, no.~1, pp. 354--366, 2023.

\bibitem{DBLP:conf/bmvc/KazakosHNZD21}
E.~Kazakos, J.~Huh, A.~Nagrani, A.~Zisserman, and D.~Damen, ``With a little
  help from my temporal context: Multimodal egocentric action recognition,'' in
  \emph{32nd British Machine Vision Conference 2021, {BMVC} 2021, Online,
  November 22-25, 2021}.\hskip 1em plus 0.5em minus 0.4em\relax {BMVA} Press,
  2021, p. 268.

\bibitem{DBLP:conf/cvpr/Zhang0Z21}
C.~Zhang, A.~Gupta, and A.~Zisserman, ``Temporal query networks for
  fine-grained video understanding,'' in \emph{Proceedings of the IEEE/CVF
  Conference on Computer Vision and Pattern Recognition}.\hskip 1em plus 0.5em
  minus 0.4em\relax Computer Vision Foundation / {IEEE}, 2021, pp. 4486--4496.

\bibitem{Zhou_2023_WACV}
J.~Zhou and Y.~Wu, ``Temporal feature enhancement dilated convolution network
  for weakly-supervised temporal action localization,'' in \emph{Proceedings of
  the IEEE/CVF Winter Conference on Applications of Computer Vision}, January
  2023, pp. 6028--6037.

\bibitem{Nag_2023_WACV}
S.~Nag, O.~Goldstein, and A.~K. Roy-Chowdhury, ``Semantics guided contrastive
  learning of transformers for zero-shot temporal activity detection,'' in
  \emph{Proceedings of the IEEE/CVF Winter Conference on Applications of
  Computer Vision (WACV)}, January 2023, pp. 6243--6253.

\bibitem{8741082}
J.~Huang, N.~Li, T.~Li, S.~Liu, and G.~Li, ``Spatial–temporal context-aware
  online action detection and prediction,'' \emph{IEEE Transactions on Circuits
  and Systems for Video Technology}, vol.~30, no.~8, pp. 2650--2662, 2020.

\bibitem{10058582}
K.~Xia, L.~Wang, Y.~Shen, S.~Zhou, G.~Hua, and W.~Tang, ``Exploring action
  centers for temporal action localization,'' \emph{IEEE Transactions on
  Multimedia}, pp. 1--13, 2023.

\bibitem{DBLP:conf/bmvc/DaiDB21}
R.~Dai, S.~Das, and F.~Br{\'{e}}mond, ``{CTRN:} class-temporal relational
  network for action detection,'' in \emph{32nd British Machine Vision
  Conference 2021, {BMVC} 2021, Online, November 22-25, 2021}.\hskip 1em plus
  0.5em minus 0.4em\relax {BMVA} Press, 2021, p. 224.

\bibitem{piergiovanni2018learning}
A.~Piergiovanni and M.~S. Ryoo, ``Learning latent super-events to detect
  multiple activities in videos,'' in \emph{Proceedings of the IEEE/CVF
  Conference on Computer Vision and Pattern Recognition}, 2018, pp. 5304--5313.

\bibitem{zhu2021semantic}
C.~Zhu, F.~Chen, U.~Ahmed, Z.~Shen, and M.~Savvides, ``Semantic relation
  reasoning for shot-stable few-shot object detection,'' in \emph{Proceedings
  of the IEEE/CVF Conference on Computer Vision and Pattern Recognition}, 2021,
  pp. 8782--8791.

\bibitem{nie2022node}
H.~Nie, R.~Wang, and X.~Chen, ``From node to graph: Joint reasoning on
  visual-semantic relational graph for zero-shot detection,'' in
  \emph{Proceedings of the IEEE/CVF Winter Conference on Applications of
  Computer Vision}, 2022, pp. 1109--1118.

\bibitem{yang2022sega}
F.~Yang, R.~Wang, and X.~Chen, ``Sega: semantic guided attention on visual
  prototype for few-shot learning,'' in \emph{Proceedings of the IEEE/CVF
  Winter Conference on Applications of Computer Vision}, 2022, pp. 1056--1066.

\bibitem{wang2022beyond}
Y.~Wang, C.~Cao, and Y.~Zhang, ``Beyond vision: A semantic reasoning enhanced
  model for gesture recognition with improved spatiotemporal capacity,'' in
  \emph{Chinese Conference on Pattern Recognition and Computer Vision
  (PRCV)}.\hskip 1em plus 0.5em minus 0.4em\relax Springer, 2022, pp. 420--434.

\bibitem{piergiovanni2019temporal}
A.~Piergiovanni and M.~Ryoo, ``Temporal gaussian mixture layer for videos,'' in
  \emph{International Conference on Machine learning}.\hskip 1em plus 0.5em
  minus 0.4em\relax PMLR, 2019, pp. 5152--5161.

\bibitem{dai2021pdan}
R.~Dai, S.~Das, L.~Minciullo, L.~Garattoni, G.~Francesca, and F.~Bremond,
  ``Pdan: Pyramid dilated attention network for action detection,'' in
  \emph{Proceedings of the IEEE/CVF Winter Conference on Applications of
  Computer Vision}, 2021, pp. 2970--2979.

\bibitem{yeung2018every}
S.~Yeung, O.~Russakovsky, N.~Jin, M.~Andriluka, G.~Mori, and L.~Fei-Fei,
  ``Every moment counts: Dense detailed labeling of actions in complex
  videos,'' \emph{International Journal of Computer Vision}, vol. 126, no.~2,
  pp. 375--389, 2018.

\bibitem{dai2022toyota}
R.~Dai, S.~Das, S.~Sharma, L.~Minciullo, L.~Garattoni, F.~Bremond, and
  G.~Francesca, ``Toyota smarthome untrimmed: Real-world untrimmed videos for
  activity detection,'' \emph{IEEE Transactions on Pattern Analysis and Machine
  Intelligence}, 2022.

\bibitem{sultani2018real}
W.~Sultani, C.~Chen, and M.~Shah, ``Real-world anomaly detection in
  surveillance videos,'' in \emph{Proceedings of the IEEE Conference on
  Computer Vision and Pattern Recognition}, 2018, pp. 6479--6488.

\bibitem{liu2019exploring}
K.~Liu and H.~Ma, ``Exploring background-bias for anomaly detection in
  surveillance videos,'' in \emph{Proceedings of the 27th ACM International
  Conference on Multimedia}, 2019, pp. 1490--1499.

\bibitem{ozturk2021adnet}
H.~{\.I}. {\"O}zt{\"u}rk and A.~B. Can, ``Adnet: Temporal anomaly detection in
  surveillance videos,'' in \emph{International Conference on Pattern
  Recognition}.\hskip 1em plus 0.5em minus 0.4em\relax Springer, 2021, pp.
  88--101.

\bibitem{DBLP:journals/tmm/ChenZZ21}
G.~Chen, C.~Zhang, and Y.~Zou, ``Afnet: Temporal locality-aware network with
  dual structure for accurate and fast action detection,'' \emph{{IEEE} Trans.
  Multim.}, vol.~23, pp. 2672--2682, 2021.

\bibitem{DBLP:conf/iccv/XuDS17}
H.~Xu, A.~Das, and K.~Saenko, ``{R-C3D:} region convolutional 3d network for
  temporal activity detection,'' in \emph{{IEEE} International Conference on
  Computer Vision, {ICCV} 2017, Venice, Italy, October 22-29, 2017}.\hskip 1em
  plus 0.5em minus 0.4em\relax {IEEE} Computer Society, 2017, pp. 5794--5803.

\bibitem{lin2019bmn}
T.~Lin, X.~Liu, X.~Li, E.~Ding, and S.~Wen, ``Bmn: Boundary-matching network
  for temporal action proposal generation,'' in \emph{Proceedings of the
  IEEE/CVF International Conference on Computer Vision}, 2019, pp. 3889--3898.

\bibitem{shou2017cdc}
Z.~Shou, J.~Chan, A.~Zareian, K.~Miyazawa, and S.-F. Chang, ``Cdc:
  Convolutional-de-convolutional networks for precise temporal action
  localization in untrimmed videos,'' in \emph{Proceedings of the IEEE/CVF
  Conference on Computer Vision and Pattern Recognition}, 2017, pp. 5734--5743.

\bibitem{DBLP:conf/cvpr/HeilbronNG16}
F.~C. Heilbron, J.~C. Niebles, and B.~Ghanem, ``Fast temporal activity
  proposals for efficient detection of human actions in untrimmed videos,'' in
  \emph{Proceedings of the IEEE/CVF Conference on Computer Vision and Pattern
  Recognition}, 2016, pp. 1914--1923.

\bibitem{DBLP:conf/eccv/EscorciaHNG16}
V.~Escorcia, F.~C. Heilbron, J.~C. Niebles, and B.~Ghanem, ``Daps: Deep action
  proposals for action understanding,'' in \emph{Computer Vision - {ECCV} 2016
  - 14th European Conference, Amsterdam, The Netherlands, October 11-14, 2016,
  Proceedings, Part {III}}, B.~Leibe, J.~Matas, N.~Sebe, and M.~Welling, Eds.

\bibitem{kahatapitiya2021coarse}
K.~Kahatapitiya and M.~S. Ryoo, ``Coarse-fine networks for temporal activity
  detection in videos,'' in \emph{Proceedings of the IEEE/CVF Conference on
  Computer Vision and Pattern Recognition}, 2021, pp. 8385--8394.

\bibitem{dai2022ms}
R.~Dai, S.~Das, K.~Kahatapitiya, M.~S. Ryoo, and F.~Bremond, ``Ms-tct:
  Multi-scale temporal convtransformer for action detection,'' in
  \emph{Proceedings of the IEEE/CVF Conference on Computer Vision and Pattern
  Recognition}, 2022, pp. 20\,041--20\,051.

\bibitem{dai2019self}
R.~Dai, L.~Minciullo, L.~Garattoni, G.~Francesca, and F.~Bremond,
  ``Self-attention temporal convolutional network for long-term daily living
  activity detection,'' in \emph{2019 16th IEEE International Conference on
  Advanced Video and Signal Based Surveillance (AVSS)}.\hskip 1em plus 0.5em
  minus 0.4em\relax IEEE, 2019, pp. 1--7.

\bibitem{lea2017temporal}
C.~Lea, M.~D. Flynn, R.~Vidal, A.~Reiter, and G.~D. Hager, ``Temporal
  convolutional networks for action segmentation and detection,'' in
  \emph{proceedings of the IEEE Conference on Computer Vision and Pattern
  Recognition}, 2017, pp. 156--165.

\bibitem{long2019gaussian}
F.~Long, T.~Yao, Z.~Qiu, X.~Tian, J.~Luo, and T.~Mei, ``Gaussian temporal
  awareness networks for action localization,'' in \emph{Proceedings of the
  IEEE/CVF Conference on Computer Vision and Pattern Recognition}, 2019, pp.
  344--353.

\bibitem{DBLP:conf/accv/ZhengHHHC20}
Y.~Zheng, R.~Huang, C.~Han, X.~Huang, and L.~Cui, ``Background learnable
  cascade for zero-shot object detection,'' in \emph{Computer Vision - {ACCV}
  2020 - 15th Asian Conference on Computer Vision, Kyoto, Japan, November 30 -
  December 4, 2020, Revised Selected Papers, Part {III}}, H.~Ishikawa, C.~Liu,
  T.~Pajdla, and J.~Shi, Eds., 2020.

\bibitem{DBLP:conf/cvpr/ZhuWS20}
P.~Zhu, H.~Wang, and V.~Saligrama, ``Don't even look once: Synthesizing
  features for zero-shot detection,'' in \emph{Proceedings of the IEEE/CVF
  Conference on Computer Vision and Pattern Recognition}.\hskip 1em plus 0.5em
  minus 0.4em\relax Computer Vision Foundation / {IEEE}, 2020, pp.
  11\,690--11\,699.

\bibitem{caba2015activitynet}
F.~Caba~Heilbron, V.~Escorcia, B.~Ghanem, and J.~Carlos~Niebles, ``Activitynet:
  A large-scale video benchmark for human activity understanding,'' in
  \emph{Proceedings of the IEEE/CVF Conference on Computer Vision and Pattern
  Recognition}, 2015, pp. 961--970.

\bibitem{THUMOS14}
Y.-G. Jiang, J.~Liu, A.~Roshan~Zamir, G.~Toderici, I.~Laptev, M.~Shah, and
  R.~Sukthankar, ``{THUMOS} challenge: Action recognition with a large number
  of classes,'' \url{http://crcv.ucf.edu/THUMOS14/}, 2014.

\bibitem{DBLP:journals/corr/LiuHLSL17}
C.~Liu, Y.~Hu, Y.~Li, S.~Song, and J.~Liu, ``{PKU-MMD:} {A} large scale
  benchmark for continuous multi-modal human action understanding,''
  \emph{CoRR}, 2017.

\bibitem{zhao2019hacs}
H.~Zhao, A.~Torralba, L.~Torresani, and Z.~Yan, ``Hacs: Human action clips and
  segments dataset for recognition and temporal localization,'' in
  \emph{Proceedings of the IEEE/CVF International Conference on Computer
  Vision}, 2019, pp. 8668--8678.

\bibitem{DBLP:journals/corr/abs-1301-3781}
T.~Mikolov, K.~Chen, G.~Corrado, and J.~Dean, ``Efficient estimation of word
  representations in vector space,'' in \emph{1st International Conference on
  Learning Representations, {ICLR} 2013, Scottsdale, Arizona, USA, May 2-4,
  2013, Workshop Track Proceedings}, Y.~Bengio and Y.~LeCun, Eds., 2013.

\bibitem{pennington2014glove}
J.~Pennington, R.~Socher, and C.~D. Manning, ``Glove: Global vectors for word
  representation,'' in \emph{Proceedings of the 2014 Conference on Empirical
  Methods in Natural Language Processing (EMNLP)}, 2014, pp. 1532--1543.

\bibitem{DBLP:conf/emnlp/WangTI21}
S.~Wang, L.~Thompson, and M.~Iyyer, ``Phrase-bert: Improved phrase embeddings
  from {BERT} with an application to corpus exploration,'' in \emph{Proceedings
  of the 2021 Conference on Empirical Methods in Natural Language Processing,
  {EMNLP} 2021, Virtual Event / Punta Cana, Dominican Republic, 7-11 November,
  2021}, M.~Moens, X.~Huang, L.~Specia, and S.~W. Yih, Eds.\hskip 1em plus
  0.5em minus 0.4em\relax Association for Computational Linguistics, 2021, pp.
  10\,837--10\,851.

\bibitem{DBLP:conf/naacl/DevlinCLT19}
J.~Devlin, M.~Chang, K.~Lee, and K.~Toutanova, ``{BERT:} pre-training of deep
  bidirectional transformers for language understanding,'' in \emph{Proceedings
  of the 2019 Conference of the North American Chapter of the Association for
  Computational Linguistics: Human Language Technologies}, J.~Burstein,
  C.~Doran, and T.~Solorio, Eds.\hskip 1em plus 0.5em minus 0.4em\relax
  Association for Computational Linguistics, 2019, pp. 4171--4186.

\bibitem{carreira2017quo}
J.~Carreira and A.~Zisserman, ``Quo vadis, action recognition? a new model and
  the kinetics dataset,'' in \emph{Proceedings of the IEEE/CVF Conference on
  Computer Vision and Pattern Recognition}, 2017, pp. 6299--6308.

\bibitem{DBLP:conf/cvpr/MahasseniT16}
B.~Mahasseni and S.~Todorovic, ``Regularizing long short term memory with 3d
  human-skeleton sequences for action recognition,'' in \emph{Proceedings of
  the IEEE/CVF Conference on Computer Vision and Pattern Recognition}.\hskip
  1em plus 0.5em minus 0.4em\relax {IEEE} Computer Society, 2016, pp.
  3054--3062.

\end{thebibliography}
 \bibliographystyle{IEEEtran}
%

\newpage

%
%
%
%

\vfill

\end{document}